\documentclass[11pt,a4paper]{article}
\usepackage[hyperref]{acl2021}
\usepackage{times}
\usepackage{latexsym}

\usepackage{microtype}

\usepackage{amssymb}
\usepackage{amsmath}
\usepackage{amsfonts}
\usepackage{graphicx}
\usepackage{fontawesome}

\usepackage{enumitem}
\usepackage{subcaption}
\usepackage{rotating,csquotes}
\usepackage{xcolor,color,colortbl}
\usepackage{multirow}
\usepackage{framed}
\usepackage{booktabs}
\usepackage{array}

\aclfinalcopy %
\def\aclpaperid{2864} %

\newcommand{\PreserveBackslash}[1]{\let\temp=\\#1\let\\=\temp}
\newcolumntype{C}[1]{>{\PreserveBackslash\centering}p{#1}}
\newcolumntype{R}[1]{>{\PreserveBackslash\raggedleft}p{#1}}
\newcolumntype{L}[1]{>{\PreserveBackslash\raggedright}p{#1}}

\newcommand{\scism}[2]{{#1}\text{e-}{#2}}

\newcommand{\loss}{\mathcal{L}}

\newcommand{\knn}{$k$NN}{}
\newcommand{\vanilla}{vanilla-{\knn}}{}
\newcommand{\Vanilla}{Vanilla-{\knn}}{}
\newcommand{\minimax}{MiniMax-{\knn}}{}
\newcommand{\nkd}[1]{vanilla-{#1}NN}{}
\newcommand{\nmm}[1]{MiniMax-{#1}NN}{}
\newcommand{\nrand}[1]{random-{#1}}{}
\newcommand{\RoB}{RoBERTa}{}
\newcommand{\robertal}{{\RoB}$_\text{Large}$}{}
\newcommand{\distRoB}{Distil{\RoB}}{}
{}
{}

\title{Not Far Away, Not So Close: Sample Efficient Nearest Neighbour Data Augmentation via MiniMax}

\author{
  Ehsan Kamalloo\thanks{\ \ Equal Contribution} $^{\:\dagger\lozenge}$\quad
  Mehdi Rezagholizadeh$^{*\S}$ \quad Peyman Passban\thanks{\ \ Work done while at Huawei Noah's Ark Lab} $^{\:\S}$ \quad Ali Ghodsi$^{\ddagger\P}$\\
  $^\lozenge$Department of Computing Science, University of Alberta \\
  $^\S$Huawei Noah's Ark Lab \\
$^\ddagger$David R. Cheriton School of Computer Science, Univeristy of Waterloo \\
$^\P$Department of Statistics and Actuarial Science, Univeristy of Waterloo \\
  \texttt{kamalloo}\texttt{@cs.ualberta.ca}
}

\date{}

\begin{document}
\maketitle
\begin{abstract}
In Natural Language Processing (NLP), finding data augmentation techniques that can produce high-quality {\em human-interpretable} examples has always been challenging.
Recently, leveraging {\knn} such that augmented examples are retrieved from large repositories of unlabelled sentences has made a step toward interpretable augmentation.
Inspired by this paradigm, we introduce {\em \minimax}, a sample efficient data augmentation strategy tailored for Knowledge Distillation (KD).
We exploit a semi-supervised approach based on KD to train a model on augmented data.
In contrast to existing {\knn} augmentation techniques that blindly incorporate all samples, our method dynamically selects a subset of augmented samples that maximizes KL-divergence between the teacher and student models.
This step aims to extract the most efficient samples to ensure our augmented data covers regions in the input space with maximum loss value.
We evaluated our technique on several text classification tasks and demonstrated that {\minimax} consistently outperforms strong baselines. Our results show that {\minimax} requires fewer augmented examples and less computation to achieve superior performance over the state-of-the-art {\knn}-based augmentation techniques.

\end{abstract}

\section{Introduction}

Knowledge distillation (KD) \cite{bucilua2006model, hinton2015distilling} has been successful in improving the performance of various NLP tasks such as language modelling~\cite{jiao-etal-2020-tinybert, sanh2019distilbert, turc2019well}, machine translation~\cite{tan2019multilingual,wu-etal-2020-skip}, natural language understanding~\cite{passban2020alp, rashid2021mate}, and multi-task learning~\cite{clark-etal-2019-bam}.
It aims to transfer the knowledge embedded in a model---called teacher---to another succedent model---called student, without compromising on accuracy \cite{furlanello2018born}.

Data plays a significant role in the success of KD. The importance of data becomes even more crucial when dealing with large teacher models~\cite{lopez2015unifying} or managing tasks with small amount of labelled data~\cite{rashid2020towards, nayak2019zero}.
The training objective of KD focuses on minimizing the discrepancy between representations of a teacher model and a student model. However, this might not be the case for regions which are not covered by training data in the input space.
Data augmentation comes into play as a natural solution for such circumstances.

Most existing data augmentation techniques are not tailored for KD as the dynamics of teacher and student models are not considered in generating augmented data. Moreover, other model-based data augmentation techniques such as adversarial approaches do not generate interpretable samples for NLP tasks~\cite{du-etal-2021-self}.  
In this work, inspired by the success of retrieval-based augmentation techniques~\cite{guu2020realm, khandelwal2019generalization,du-etal-2021-self, kassner-schutze-2020-bert}, 
we propose {\em {\minimax}}, an interpretable data augmentation methodology. Our technique is interleaved with KD training to generate realistically-looking training points. For this purpose, we use a massive external respostiory of unlabelled sentences.
In contrast to previous {\knn} augmentation techniques which naively extract and incorporate $k$ samples,
we propose a minimax approach to adapt {\knn} augmentation to KD and select our augmented samples more efficiently.

Experimental results show that our technique requires significantly fewer samples, reaches the state-of-the-art {\knn} augmentation technique~\cite{du-etal-2021-self}, and improves generalization to unseen data.\footnote{Source code is available at \url{https://github.com/ehsk/Minimax-kNN}}

Our key contributions can be summarized as follows:
\begin{itemize}
    \item We tailor {\knn}-based data augmentation for KD via MiniMax to select more impactful augmented samples for training.
    \item We significantly improve sample efficiency of {\knn}-based data augmentation.
    \item We conduct extensive experiments to evaluate our proposed method and manifest that we can maintain the test performance with training on only influential augmented examples.
\end{itemize}

\section{Background}

\subsection{Data Augmentation in KD}
KD \cite{hinton2015distilling} is a training method that incorporates the knowledge of a teacher network
in training a student network. 
The teacher can be trained on the same dataset as the student and often provides a suitable approximation of the underlying distribution of data.
The training loss of the student using KD is formulated as in Eq.~\eqref{eq:KD}. 

\begin{equation}
\begin{split}
    & \mathcal{L} =(1-\lambda)\mathcal{L}_{CE} + \lambda \mathcal{L}_{KD} \\
    & \mathcal{L}_{CE} = {CE}\Big(y,\sigma(z_s(x)\Big)  \\
    & \mathcal{L}_{KD} = \mathcal{T}^2 KL\Big(\sigma(\frac{z_t(x)}{\mathcal{T}}), \sigma(\frac{z_s(x)}{\mathcal{T}})\Big)
\end{split}
\label{eq:KD}
\end{equation}
where $z_s$ and $z_t$ refer to the logits of the student and teacher networks, $\sigma(.)$ is the softmax prediction, $CE$ and $KL$ refer to cross entropy and KL-divergence loss, respectively. $\lambda$ is a hyper-parameter which controls the contribution of the KD loss with respect to the original cross entropy loss, and $\mathcal{T}$ is the temperature parameter which determines the smoothness of the output probability. 

Although KD has been shown to be successful in model compression \cite{bucilua2006model} and improving the performance of neural networks \cite{furlanello2018born}, the core prerequisites for effective KD are often overlooked. \citet{lopez2015unifying} give a good insight about these conditions using the VC-dimension analysis:
\begin{equation}
O(\frac{|\mathcal{F}_s|_c+|\mathcal{F}_t|_c}{n^{\alpha}}) + \varepsilon_{t}+\varepsilon_{l} \leq O(\frac{|\mathcal{F}_s|_c}{\sqrt{n}}) + \varepsilon_{s}
\label{eq:vap}
\end{equation} 
where 
$\mathcal{F}_s$ and $\mathcal{F}_t$ are the function classes corresponding to the teacher and student;  $|.|_c$ is a function class capacity measure; $O(.)$ is the estimation error of training the learner; $\varepsilon_s$ is the approximation error of the best estimator function belonging to the $\mathcal{F}_s$ class with respect to the underlying function; $\varepsilon_t$ is a similar approximation error for the teacher with respect to the underlying function; $\varepsilon_l$ is the approximation error of the best student function with respect to the teacher function; $n$ is the number of training samples, and $\frac{1}{2}\leq \alpha \leq 1$ is a parameter related to the difficulty of the problem.

\begin{figure}
  \centering
  \includegraphics[width=\linewidth]{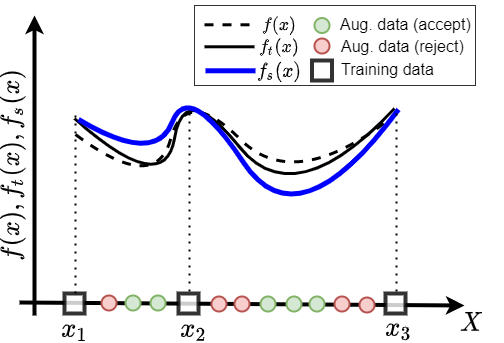}
  \caption{\label{fig:data} Data sparsity problem in KD; $f$, $f_t$, and $f_s$ are representing the underlying function, teacher, and student outputs respectively. We show 10 augmented samples around $x_2$ with small circles on the X-axis. The green circles show the augmented samples which are selected by our {\minimax} because these points correspond to maximum divergence regions of the teacher and student networks. The red circles are rejected augmented samples. }
\end{figure}

According to Eq.~\eqref{eq:vap}, it is clear that when the capacity of the teacher is large or when the number of the training samples is small, training with KD can be less beneficial. Figure~\ref{fig:data} illustrates this problem through a synthetic example that KD loss forces the student to follow the teacher on training samples but there is no guarantee for such phenomenon to happen in regions in the input space that are not covered by training data.
Therefore, the chance of a mismatch between two networks would be higher if training data is sparse or when there is a large gap between two networks.

Data augmentation can be considered as a remedy for this problem. To the best of our knowledge, most existing techniques are not sample efficient and blindly consider all generated samples in their training. As illustrated in Figure~\ref{fig:data}, different augmented samples might have different contribution to the final teacher/student loss.  
Moreover, these augmentation techniques are not tailored for KD. Our {\minimax} solution addresses these two problems.

\subsection{Nearest Neighbour Data Augmentation}
\label{sec:knn}

The {\knn} augmentation strategy consists of two main stages: (a) a paraphrastic nearest neighbour retrieval engine, and (b) a training method using augmented samples.

Initially, training examples are queried over a large sentence repository using a general-purpose paraphrastic encoder. The aim of this stage is to find interpretable unannotated augmented samples that are semantically close to training data. For this purpose, we use one of the sentence repositories from SentAugment \cite{du-etal-2021-self}, comprising 100M sentences collected from Common Crawl. We also employ the same paraphrastic sentence encoder, namely SASE, introduced in SentAugment. SASE is an XLM model \cite{lample2019cross}, fine-tuned on a number of well-known paraphrase datasets using a triplet loss to maximize the cosine similarity between representations of paraphrases. The similarity between a pair of sentence representations obtained from SASE can be adopted for unsupervised semantic similarity. \citet{du-etal-2021-self} show that SASE achieves high correlation (0.73 on average) with human judgment on several STS benchmarks. Consequently, the {\knn} operation can be summarized as follows: Suppose a dataset $\left\{x_i, y_i\right\}_{i=1}^{N}$ where $x_i$ and $y_i$ denote an example and its corresponding label respectively. Given a large sentence repository $\mathcal{R}$ encoded using SASE, {\knn} is determined via top $k$ sentences with respect to $\cos(\text{SASE}(x_i), \text{SASE}(s_j))$ where $s_j \in \mathcal{R}$.

Next, in step (b), a model is trained on the original data by minimizing $\loss_{CE}$ from Eq.~\eqref{eq:KD}. The trained model learns task-specific knowledge that is further useful in finding relevant augmented examples. To this end, retrieved examples that are close to original examples within the teacher's space are retained to form augmented data.
Augmented examples are subsequently incorporated into training via KD. 
A student model is then distilled from the teacher by leveraging teacher's soft labels on the combination of original data and augmented samples. In particular, for original examples $\{x_i, y_i\}$, $\loss$ from Eq. \eqref{eq:KD} is minimized during training, whereas for the augmented examples, we only minimize $\loss_{KD}$.

\section{Related Work}
\subsection{KD in Tandem with Data Augmentation}
Adaptive data augmentation can strengthen the capacity of the teacher in transferring knowledge to the student during distillation \cite{fu2020role}. Numerous studies \cite{chen2020big, xie2020self} have applied KD for self-training in image classification tasks. In NLP, however, generating semantically plausible examples that can be easily inspected by humans is more challenging. In TinyBERT \cite{jiao-etal-2020-tinybert}, a contextual augmentation method is used along with KD, but such augmentation does not take the advantages of teacher or student's knowledge. 
A recent paradigm that heavily relies on data augmentation is zero-shot KD \cite{nayak2019zero, rashid2020towards}. In contrast, we explore the interpretability of augmentation in KD, which distinguishes our approach from the literature.

\subsection{Data Augmentation in NLP}
Word-level methods \cite{zhang2015character, xie2017data, wei-zou-2019-eda} are heuristic based and do not necessarily yield natural sentences. More recently, contextual augmentations \cite{kobayashi2018contextual, yi2021reweighting} that substitute words for other words, is shown effective in text classification. However, these approaches do not produce diverse syntactic forms. Similarly, inspired by denoising auto-encoders, augmented examples can be sampled from the reconstruction distribution of corrupted sentences via Masked Language Modelling \cite{ng-etal-2020-ssmba}.
Back-translation \cite{sennrich2016improving} is also another strategy to obtain augmented data \cite{yu2018qanet, xie2020unsupervised, chen2020semi, qu2021coda}.

Another line of work that mainly targets model robustness is to create new data or counterfactual examples via human-in-the-loop perturbations \cite{kaushik2019learning, khashabi-etal-2020-bang, jin2020bert}. Nonetheless, these strategies are task-specific and not scalable to generate data at massive scale. Besides, our method diverges from these studies in that we intend to build a semi-supervised system with minimal human intervention.

Several models \cite{miyato2017adversarial, zhu2019freelb, jiang-etal-2020-smart, cheng-etal-2020-advaug, qu2021coda} leveraged adversarial training for data augmentation.
These methods manipulate the input embedding space to construct synthetic examples. 
Neighbourhoods around training instances in the embedding space cannot be translated back to text and thus are not interpretable. Although we advocate for interpretable data augmentation, we do not compete with these techniques and in fact, gradient-based augmentation is complementary to our method.

Finally, {\knn}, a non-parametric search algorithm that probes an external data source to find nearest neighbours is employed in several NLP tasks such as language modelling \cite{khandelwal2019generalization}, machine translation \cite{khandelwal2021nearest}, cloze question answering \cite{kassner-schutze-2020-bert}, and open-domain question answering \cite{lewis2020retrieval}. {\knn} offers access to explicit memory that can retrieve factual knowledge from a data store. {\knn} is highly interpretable as knowledge is stored in raw text, an easy format for humans to understand. Recently, SentAugment \cite{du-etal-2021-self} introduced a semi-supervised strategy with unlabelled sentences. It retrieves augmented samples from a universal data store using {\knn}. Our proposed strategy is in line with SentAugment at heart, but different in leveraging the augmented examples during training. We focus on sample efficiency and show that we can reduce the size of the augmented data---e.g., by 60\% in sentiment classification as reported in Section~\ref{sec:fewshot}---while reaching a competitive performance.

\begin{figure}
  \centering
  \includegraphics[width=\linewidth]{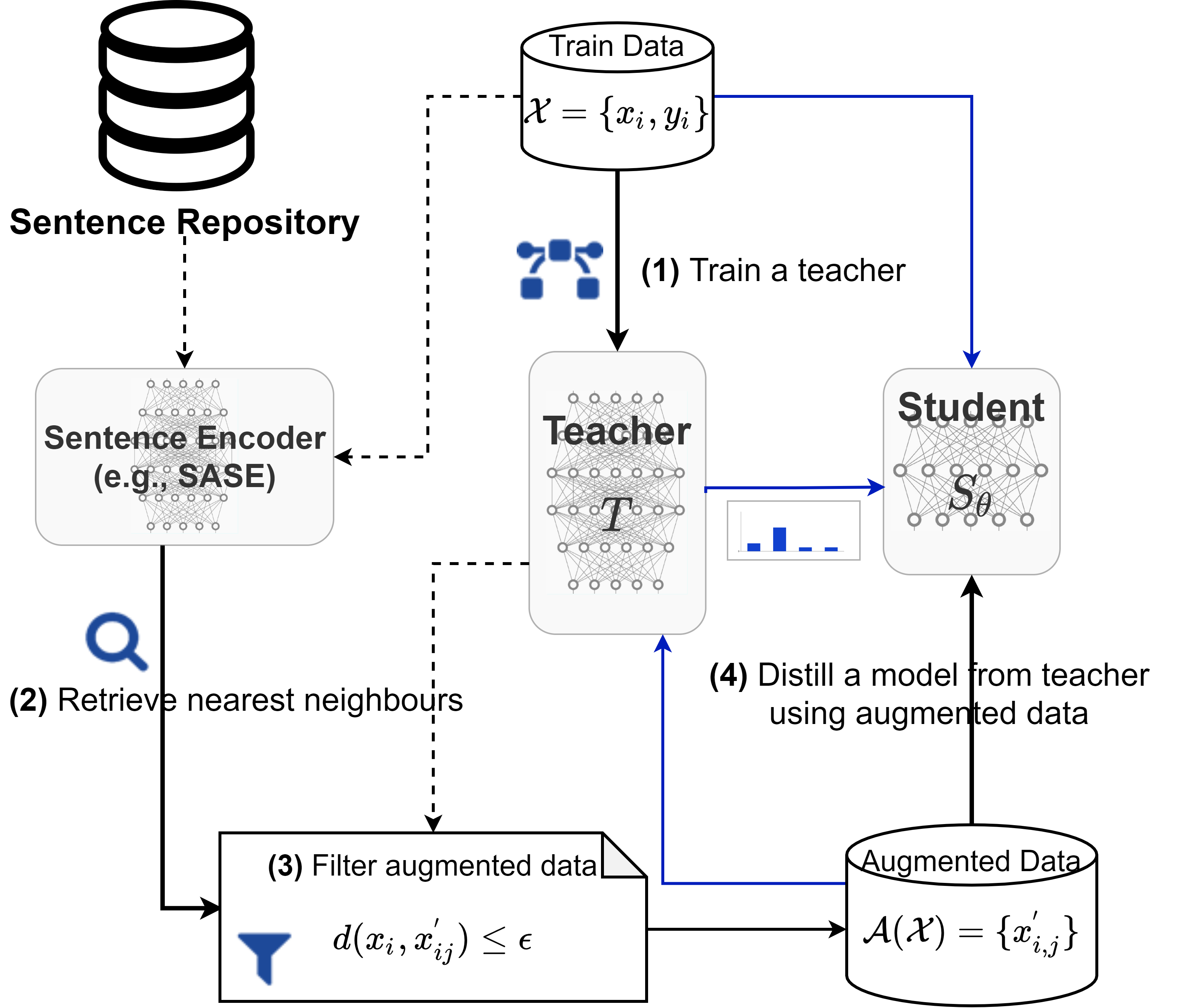}
  \caption{\label{fig:overview} A schematic view of {\minimax}}
\end{figure}

\section{{\minimax} Data Augmentation for KD}
\label{sec:minimax}

{Inspired by \citet{volpi2018generalizing} and \citet{madry2017towards}, we apply minimax framework to tailor a sample efficient {\knn} data augmentation for KD.}  
{Minimizing the maximum expected risk is used in adversarial training \cite{volpi2018generalizing} and it is shown to have guaranteed good performance on distributions ($P$) within a particular distance ($\rho$) from a source distribution ($P_0$):}
\begin{equation}
\min_\theta \sup_{\mathcal{D}(P,P_0)\leq \rho} \mathbf{E} [l(x',y';\theta)] 
\label{minimax}
\end{equation}
where $\mathcal{D}$ is a notion of distance between distributions, $l$ refers to the loss function, $\theta$ represents the parameters of the estimator model, and in our framework, $(x',y')$ are augmented data samples.   

Let us define the set of {\knn} augmented samples corresponding to the training sample $x_i \in \mathcal{X}$ from the training set, $\mathcal{X}$,  to be $\mathcal{A}(x_i)=\{x'_{i1}, x'_{i2}, ..., x'_{ik}\}$. 
In the maximization phase, we define the loss $l(x',y';\theta)=\text{KL}\big(T(x'),S(x';\theta)\big)$ between the softmax output of the teacher $T(x')$ and that of the student network $S(x';\theta)$ with trainable parameters $\theta$, with respect to the given augmented samples. Note that the augmented samples are unlabelled in the maximization phase. Then, we sort the augmented samples based on their loss value and form our {\minimax} augmentation set $\bar{\mathcal{A}}(x_i)$ by selecting the top $n$ out of the $k$ samples in $\mathcal{A}(x_i)$. $n$ is a hyper-parameter in our method that determines the sample efficiency of {\minimax}. 
In order to enforce $\mathcal{D}(P,P_0)\leq \rho$ on the distance between the two distributions in Eq.~\eqref{minimax}, in our {\knn} search, we set a maximum radial semantic distance $\epsilon$ between the sentence representation of accepted augmented samples in $\mathcal{\bar{A}}(x_i)$ and the sentence representation of their corresponding input $x_i$ based on the angular distance metric: 
\begin{equation}
d(x_i, x'_{ij})=\frac{1}{\pi} \cos^{-1} \frac{<h_{cls}^t(x_i).h_{cls}^t(x'_{ij})>}{\|h_{cls}^t(x_i)\|\|h_{cls}^t(x'_{ij})\|} \leq \epsilon
\label{epsilon}
\end{equation}
where $h_{cls}^t$ refers to the teacher's last layer hidden representation of the \texttt{[CLS]} token, and $<\!\cdot\!>$ denotes the dot product of two vectors. The discussion on how to adjust $\epsilon$ is given in Section~\ref{minimax_epsilon}.  

In summary, our technique equips {\knn} augmentation with minimax to improve its sample efficiency. In contrast to adversarial data augmentation methods,
our approach uses the minimax loss for selecting augmented samples. The overall structure of our augmentation strategy is visualized in Figure~\ref{fig:overview}. We essentially follow three steps in each iteration during training:

\begin{itemize}[noitemsep]
    \item[\textbf{(1)}] We construct teacher logits and student logits for augmented samples to measure KL-divergence between the two models.
    \item[\textbf{(2)}] Out of all {\knn} samples, $n$ samples with highest KL-divergence will be selected.
    \item[\textbf{(3)}] KD loss is minimized for training data and selected augmented samples. 
\end{itemize}

Our experiments reveal that this modification to KD underscores sample efficiency while retaining the test performance.

\subsection{FLOPs Analysis of {\minimax}}
\label{sec:flops}
Minimax computations in {\minimax} incur additional overhead costs during training, but how much precisely do minimax operations curtail the runtime performance? To answer this question, we analyze the logical compute complexity of our algorithm in terms of floating point operations (FLOPs) because it can be measured regardless of hardware considerations \cite{clark2020electra}.

To this end, we compare FLOPs corresponding to each augmented example from {\minimax} with vanilla {\knn} within an epoch. Suppose a forward pass and a backward pass for one batch takes $F$ and $B$ FLOPs, respectively. The number of matrix operations between the forward pass and the backward pass is not considerably different and hence, $F \approx B$ \cite{clark2020electra}.
For simplicity, we assume batch size is 1. Considering $k_1$ is the number of retrieved NNs, vanilla {\knn} requires $k_1F + k_1B$ additional FLOPs per epoch.

On the other hand, {\minimax} selects $n$ neighbours from $k_2$ retrieved nearest neighbours---i.e., $n < k_2$. The algorithm first takes the logits of all $k_2$ neighbours to compute KL-divergence vectors, which needs $k_2F$ FLOPs, similar to vanilla {\knn}. The extra operations of {\minimax} occur in the maximization step in which top $n$ neighbours are determined with respect to their KL-divergence values. This operation can be carried out by sorting the KL-divergence vector, which costs $S$ FLOPs.
Note that $S \ll F$ because obtaining an output from a deep neural network model is far more costly than a sorting operation. The backward pass is then computed only for the $n$ selected neighbours. Accordingly, the overall FLOPs for {\minimax} is $k_2F + S + nB$.
The difference between the FLOPs is:
\[
\begin{split}
\Delta_{\text{FLOPs}} &= \text{FLOPs}_{\text{{\vanilla}}} - \text{FLOPs}_{\text{\minimax}} \\
&= (k_1 - k_2)F + (k_1 - n)B + S
\end{split}
\]
Given that $B$ can be approximated with $F$ (as mentioned earlier) and $S \ll F$:
\[
\Delta_{\text{FLOPs}} = (2k_1 - k_2 - n)F
\]
Thus, as long as $k_2 + n < 2k_1$, {\minimax} is more efficient than vanilla {\knn}. In experiments, we illustrate that how {\minimax} surpasses {\vanilla} while satisfying the FLOPs condition.

\begin{table}[t]
\centering
\small
 \begin{tabular}{l c c c c c} 
 \hline
 \footnotesize{\textbf{Dataset}} & \footnotesize{\textbf{\#class.}} & \footnotesize{\textbf{\#train}} & \footnotesize{\textbf{\#dev}} & \footnotesize{\textbf{\#test}} & \scriptsize{\textbf{avg. \#tokens}} \\
 \hline
 \small{SST-2} & \small{2} & \footnotesize{67.3K} & \small{872} & \small{1.8K}	& \small{12.4} \\
 \small{SST-5} & \small{5} & \small{8.5K} & \small{1.1K} & \small{2.2K}	& \small{22.6}	 \\
 \small{TREC} & \small{6} & \small{5K} & \small{500} & \small{500}  & \small{11.4} \\
 \small{CR} & \small{2} & \small{2.5K} & \small{640}  & \small{642}  & \small{21.4} \\
 \small{IMP} & \small{2} & \small{3.9K} & \small{2.2K} & \small{2.6K}  & \small{50.0} \\
 \hline
\end{tabular}
\caption{\label{tab:datasets} Downstream tasks used for evaluation}
\end{table}

\section{Experiments}
\subsection{Datasets}
We evaluate {\minimax} on five datasets:
SST-2 and SST-5 \cite{socher2013recursive} for sentiment analysis, TREC \cite{li2002learning} for question type classification, CR \cite{hu2004mining} for product review classification, and Impremium's hate-speech detection dataset (IMP)\footnote{\url{https://www.kaggle.com/c/detecting-insults-in-social-commentary/}}.
Information related to all datasets is summarized in Table~\ref{tab:datasets}.

\begin{table*}[t]
\centering
 \begin{tabular}{l c c c c c} 
 \hline
 \textbf{Model} & \textbf{SST-5} & \textbf{SST-2} & \textbf{TREC} & {\bf CR} & {\bf IMP} \\
 \hline
 {\robertal} (Teacher) & 57.6 & 96.2 & 98.0 & 94.1 & 90.0 \\
 \hline
 {\distRoB} & 52.9 & 93.5 & 96.0 & \textbf{92.1} & 86.8	 \\
 {\distRoB} + KD & 53.2 & 93.6 & 96.6 & \textbf{92.1} & 87.7 \\
 \hline
 {\distRoB} + \nkd{8} & 55.2 & 94.7 & 97.0 & 91.3 & 88.4 \\
 \small{\space\space\space\space\textsc{Aug. Size} (\#forward / \#backward pass)} & \small{8x / 8x} & \small{8x / 8x} & \small{8x / 8x} & \small{8x / 8x} & \small{8x / 8x} \\
 \hline
 {\distRoB} + \nmm{8}$^*$ & {\bf 55.4} & {\bf 95.2} & \textbf{97.6} & 91.6 & \textbf{88.6} \\
 \small{\space\space\space\space\textsc{Aug. Size} (\#forward / \#backward pass)} & \small{5x / 4x} & \small{7x / 4x} & \small{8x / 4x}  & \small{8x / 2x} & \small{8x / 1x} \\
 \hline
\end{tabular}
\caption{\label{tab:knnkd} Test accuracy ($\uparrow$) on the downstream tasks ($^*$denotes our approach and {\bf bold} numbers indicate the best result---excluding the teacher---for each task).}
\end{table*}

\subsection{Experimental Setup} \label{ssec:setup}
We adopt the publicly available pre-trained {\robertal} \cite{liu2019roberta} and {\distRoB} \cite{sanh2019distilbert}---using the Huggingface Transformers library \cite{wolf-etal-2020-transformers} and the Pytorch Lightning library\footnote{\url{https://github.com/PyTorchLightning/pytorch-lightning}}---for evaluating our approach. For KD, {\robertal} is selected as the teacher. 
For training, we adhere to findings in \citet{mosbach2020stability} and \citet{zhang2021revisiting} to circumvent the fine-tuning instability problem by training for longer iterations---i.e., 100 epochs---with early stopping and use Adam optimizer with bias correction. The model is evaluated on the development data at the end of each epoch and the best performing model is chosen for testing. Our learning rate schedule follows a
linear decay scheduler with a warm-up on $\{$10\%, 20\%$\}$ of the total number of training steps. The learning rate is tuned for each task separately out of $\{$\scism{1}{5}, \scism{2}{5}, \scism{3}{5}$\}$, and the batch size is chosen from [16, 128] depending on the dataset size.
For KD hyperparameters, we use grid search to choose the best $\lambda$ and $\mathcal{T}$ from $\{$0.3, 0.4, 0.5, 0.6$\}$ and $\{$5, 10, 12, 20$\}$.
We also schedule augmentation to start after a certain number of epochs in the training. On SST-5, SST-2, and TREC, augmentation takes effect at epochs 8, 6, and 6, respectively, whereas on IMP, and CR, augmentation starts at the beginning of training.
All experiments were conducted on two Nvidia Tesla V100 GPUs.

\paragraph{Few-shot learning setup}
We follow \citet{du-etal-2021-self} to setup the environment for few-shot learning experiments. In particular, we sample 2 training subsets with replacement from the original training set for each task. Each subset is balanced and consists of 20 examples per label. The development set is reduced to 200 examples for all tasks except CR in which we keep all of the original set. The label distribution is retained in the reduced development data. Evaluation is conducted on the actual test dataset. To obtain reliable results, we repeat training with 10 different seeds on each sampled dataset and report the average across all runs---i.e., 20 runs per task. Few-shot experiments were run on a single Nvidia Tesla V100 GPU.

\subsection{{\minimax} Results}
\label{minimax_epsilon}
First, we investigate the impact of {\knn} data augmentation at test time and compare {\minimax} with vanilla {\knn} data augmentation. To this end, we train a {\robertal} as teacher on the original data. Then, we distill a small size student based on {\distRoB} from the teacher using the augmented data and the original data.

\begin{table}[t]
\centering
\small
 \begin{tabular}{l c c c c c} 
 \hline
 \small{\textbf{Model}} & \small{\textbf{SST-5}} & \small{\textbf{SST-2}} & \small{\textbf{TREC}} & \small{{\bf CR}} & \small{{\bf IMP}} \\ 
 \hline
 \small{\nkd{8}} & \small{55.2} & \small{94.7} & \small{97.0} & \small{91.3} & \small{88.4} \\
 \hline
 \small{$n=1$} & \small{55.4} & \small{94.4} & \small{96.4} & \small{91.4} & \textbf{\small{88.6}} \\
 \small{$n=2$} & \small{54.6} & \small{95.0} & \small{96.4} & \small{91.6} & \small{88.5} \\
\small{$n=4$} & \small{55.4} & \small{\textbf{95.2}} & \small{\textbf{97.6}} & \small{91.6} & \small{87.4} \\
 \small{$n=6$} & \small{{\bf 55.6}} & \small{94.4} & \small{96.6} & \textbf{\small{91.8}} & \small{87.9} \\
 \hline
\end{tabular}
\caption{\label{tab:varyN} Test accuracy ($\uparrow$) of {\distRoB} on the downstream tasks varying the number of selected NNs ($n$) in {\nmm{8}} ({\bf bold} numbers indicate the best result for each task).}
\end{table}

\begin{table*}[t]
\centering
\small
 \begin{tabular}{l | c | c | c c | c c | c c} 
 \hline
 \multirow{2}{*}{\small{\textbf{Task}}} & \multirow{2}{*}{\small{\textbf{KD}}} & \small{$k=1$} & \multicolumn{2}{c|}{\small{$k=2$}} & \multicolumn{2}{c|}{\small{$k=4$}} & \multicolumn{2}{c}{\small{$k=8$}} \\
  & & \small{\textbf{vanilla}} & \small{\textbf{vanilla}} & \small{\textbf{MiniMax}} & \small{\textbf{vanilla}} & \small{\textbf{MiniMax}} & \small{\textbf{vanilla}} & \small{\textbf{MiniMax}} \\
 \hline
 \small{SST-5} & \small{53.2} & \small{53.9} & \small{52.0} & \small{52.5} & \small{54.7} & \small{55.0} & \small{\underline{55.2}} & \small{\textbf{55.4}} \\
 \small{SST-2} & \small{93.6} & \small{93.7} & \small{\underline{94.7}} & \small{94.2} & \small{94.6} & \small{93.8} & \small{\underline{94.7}} & \small{\textbf{95.2}} \\
\small{TREC} & \small{96.6} & \small{96.2} & \small{96.4} & \small{96.6} & \small{96.8} & \small{96.8} & \small{\underline{97.0}} & \small{\textbf{97.4}} \\
\small{CR} & \small{92.1} & \small{91.9} & \small{92.1} & \small{\underline{92.2}} & \small{\textbf{92.4}} & \small{91.6} & \small{91.3} & \small{91.9} \\
\small{IMP} & \small{87.7} & \small{87.2} & \small{87.1} & \small{87.6} & \small{86.0} & \small{87.8} & \small{\underline{88.4}} & \small{\textbf{88.6}} \\
 \hline
\end{tabular}
\caption{\label{tab:varyK} Test accuracy ($\uparrow$) of {\distRoB} on the downstream tasks varying the number of nearest neighbours ($k$). \textbf{KD} refers to knowledge distillation with no data augmentation. For MiniMax, $n$ is equal to half of $k$ neighbours  for $k=2,4$ and when $k=8$, $n$ is selected as in Table~\ref{tab:knnkd} ({\bf bold} and \underline{underline} indicate best and second best results per task).}
\end{table*}

\begin{table*}[t]
\centering
 \begin{tabular}{l c c c c c c c c c c} 
 \hline
 \textbf{Model} & \small{\textbf{SST-5}} & & \small{\textbf{SST-2}} & & \small{\textbf{TREC}} & & \small{{\bf CR}} & & \small{{\bf IMP}} & \\
 \hline
 \nkd{8} & \small{162.1} & & \small{484.5} & & \small{78.4} & & \small{43.7} & & \small{99.4} & \\
 \nmm{8}$^*$ & \small{158.8} & \small{$2\%\downarrow$} & \small{634.4} & \small{$31\%\uparrow$} & \small{101.1} & \small{$29\%\uparrow$} & \small{30.8} & \small{$30\%\downarrow$} & \small{38.6} & \small{$61\%\downarrow$} \\
 \hline
\end{tabular}
\caption{\label{tab:runtime} Training time (in seconds) for one epoch ($\downarrow$), averaged across epochs during training, on the downstream tasks along with the percent of reduction compared to \nkd{8}. \nmm{8} and \nkd{8} refer to the models we used for Table~\ref{tab:knnkd} ($^*$denotes our approach).}
\end{table*}

In Table~\ref{tab:knnkd}, we report the performance of {\minimax} as well as the {\vanilla} on the downstream tasks. In this experiment, the number of nearest neighbours ($k$) is set to 8 and for {\minimax}, we empirically select the minimum number of augmented examples ($n$) out of 8-NNs such that {\minimax} exceeds {\vanilla}. We observe that using KD alone leads to a marginal improvement on all tasks. Adding more data results in further improvements but comes at the expense of substantially longer training time. On the other hand, {\minimax} reduces the cost of training as it learns through less than half of the NNs and yet, consistently outperforms {\vanilla}.

\paragraph{Varying the number of selected examples ($n$) in {\minimax}} We explore the number of selected augmentations by varying $n$ $\in \{1, 2, 4, 6\}$ for 8-NNs on the downstream tasks. Results are reported in Table~\ref{tab:varyN}. Interestingly, picking $n$ as small as either 1 or 2 results in superior performance of {\minimax}, compared to {\vanilla}, on all tasks. In TREC, and SST-2, the sweet spot is $n=4$. In SST-5, and CR, {\minimax} performs better as $n$ grows. On the contrary, in IMP, accuracy declines by increasing $n$. 

\paragraph{Varying the number of nearest neighbours ($k$)} 
In order to investigate the optimal number of NNs, 
we assess the effect of $k$ on the downstream tasks.
The Results are reported in Table~\ref{tab:varyK}. We observe that more data sometimes makes the training noisy and as a result, performance deteriorates---e.g., $k=2$ in SST-5 and IMP. Nonetheless, when the augmentation size is sufficiently large, test results improve---i.e., $k=8$ in all datasets except CR. Apart from three cases---i.e., $k=2,4$ in SST-2, and $k=4$ in CR---{\minimax} is superior to {\vanilla} by incorporating roughly 50\% fewer examples.

\begin{figure}[t]
  \centering
  \includegraphics[width=.94\linewidth]{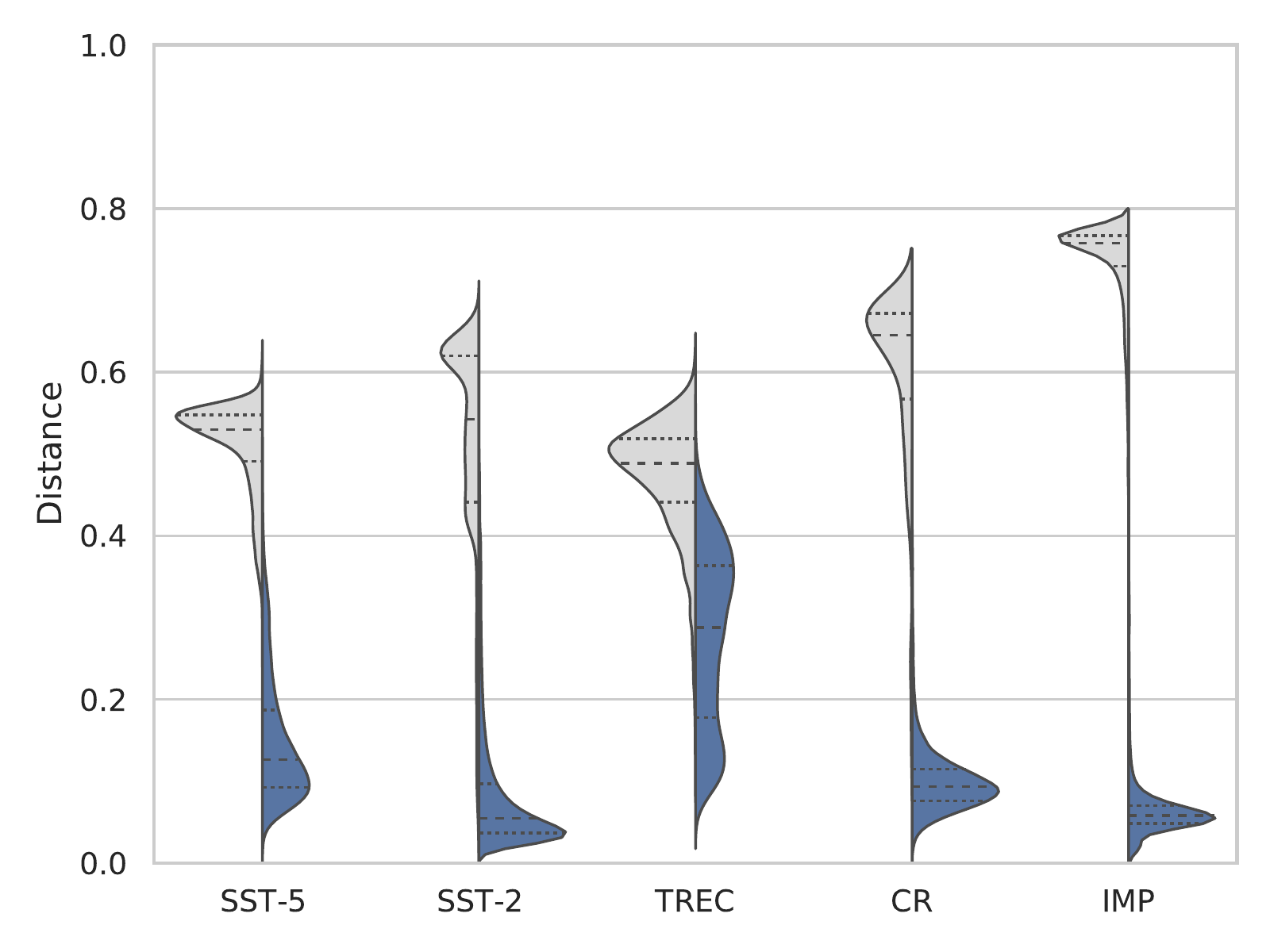}
  \caption{\label{fig:distance_distrib} Distance distribution of augmented examples for each dataset (Left: {\em mismatched labels} / Right: {\em matched labels})}
\end{figure}

\paragraph{Adjusting the maximum radial distance ($\epsilon$) in {\minimax}}
We plot the distance distribution of augmented data for two cases: (a) when the teacher predicts the same label as the original examples for augmented ones ({\em matched labels}) (b) when the predicted label for augmented examples do not match that of original examples ({\em mismatched labels}). Figure~\ref{fig:distance_distrib} illustrates a clear distinction between these two groups. Considering these insights, we find an empirical heuristic to set $\epsilon$. When the overlap between groups is infinitesimal,
we tune $\epsilon$ in the vicinity of the maximum distance of {\em matched labels}.
The rationale here is to avoid altering the skewness of the original label distribution. Throughout our experiments, $\epsilon$ is set to 0.22, and 0.4 for SST-5, and SST-2, respectively. However, we find $\epsilon=\infty$ works best on CR, IMP, and TREC.

\begin{table*}[t]
\centering
\small
 \begin{tabular}{l c c c c c} 
 \hline
 \small{\textbf{Model}} & \small{\textbf{SST-5}} & \small{\textbf{SST-2}} & \small{\textbf{TREC}} & \small{{\bf CR}} & \small{{\bf IMP}}\\
 \small{\space\space\space\space\textsc{Size}} & \small{100} & \small{40} & \small{120}  & \small{40} & \small{40} \\
 \hline
 \small{SentAugment \cite{du-etal-2021-self}} & \small{44.4 $\pm$ 1.0} & \small{86.7 $\pm$ 2.3} & \small{92.1 $\pm$ 2.4}  & \small{89.7 $\pm$ 2.0} & \small{81.9 $\pm$ 1.4} \\
 \small{\space\space\space\space\textsc{Aug. Size}} & \small{1000} & \small{1000} & \small{1000}  & \small{1000} & \small{1000} \\
 \hline
 \small{{\robertal} (Teacher)} & \small{43.9 $\pm$ 2.5} & \small{81.1 $\pm$ 2.5} & \small{89.9 $\pm$ 3.5}  & \small{83.7 $\pm$ 2.7} & \small{75.5 $\pm$ 5.5} \\
 \small{{\robertal} + KD} & \small{44.8 $\pm$ 2.5} & \small{82.3 $\pm$ 4.4} & \small{91.9 $\pm$ 2.1} & \small{83.9 $\pm$ 5.3} & \small{81.4 $\pm$ 1.9} \\
 \small{{\robertal} + \nkd{10}} & \small{45.5 $\pm$ 2.3} & \small{85.5 $\pm$ 2.9} & \small{91.6 $\pm$ 1.9} & \small{88.0 $\pm$ 1.5} & \small{81.5 $\pm$ 3.3} \\
 \small{{\robertal} + \nmm{10} ($n=6$)}$^*$ & \small{46.8 $\pm$ 1.4} & \small{86.5 $\pm$ 1.8} & \small{91.8 $\pm$ 1.0} & \small{88.3 $\pm$ 1.5} & \small{81.7 $\pm$ 2.3} \\
 \small{\space\space\space\space\textsc{Aug. Size} (\#forward / \#backward pass)} & \small{1030 / 700} & \small{380 / 280} & \small{835 / 720}  & \small{400 / 280} & \small{360 / 280} \\
 \hline
\end{tabular}
\caption{\label{tab:fewshot} Few-shot learning results of {\minimax} on the downstream tasks ($^*$denotes our approach). Compared to SentAugment, our proposed approach achieves competitive performance, but with the use of fewer augmented examples.}
\end{table*}

\subsection{Runtime Efficiency}
\label{sec:runtime}

In \S\ref{sec:flops}, we showed that {\minimax} is computationally more efficient than {\vanilla} when $k_2 + n < 2k_1$. Given the number of nearest neighbours is identical ($k_1 = k_2$) in our experiments, any choice of $n$ makes {\minimax} more efficient than {\vanilla} in theory. However, in our implementation of MiniMax, we feed selected examples again to the student, thereby triggering a redundant forward pass\footnote{All augmented examples are initially fed to the student within a PyTorch \texttt{no\_grad} block. Since we want to backpropagate through only selected examples, they should be fed again to the student.}. Although this change reduces the efficiency of {\minimax} in practice, it significantly simplifies the implementation. Thus, the above condition evolves to $k_2 + 2n < 2k_1$ in our experiments. Nonetheless, in Table~\ref{tab:knnkd}, this new efficiency constraint still holds on all tasks except SST-2, and TREC.  
To calculate the exact amount of speed-up, we measure the average training time corresponding to one epoch for each task.
The results are outlined in Table~\ref{tab:runtime}. On IMP, {\minimax} saves more than 60\% of training time and on CR, {\minimax} brings almost 30\% speed-up. Also, {\minimax} is slightly faster than {\vanilla} on SST-5. However, {\minimax} trains around 30\% slower on SST-2 and TREC.

\begin{table}[t]
\centering
\small
 \begin{tabular}{l l c c } 
 \hline
 \small{\#} & \small{\textbf{Model}} & \small{\textbf{SST-5}} & \small{$\Delta$} \\
 \hline
 \small{1} & \small{{\distRoB}} & \small{52.9} & \small{-}  \\
 \small{2} & \small{ \space + KD} & \small{53.2} & \small{+0.3} \\
 \hline
 \small{3} & \small{ \space + KD + \nrand{8}} & \small{54.4} & \small{+1.2} \\
 \small{4} & \small{ \space + KD + \nrand{8} + reranked} & \small{54.2} & \small{-0.2} \\
 \small{5} & \small{ \space + KD + \nrand{8} + reranked + $\epsilon$} & \small{52.3} & \small{-1.9} \\
 \hline
 \small{6} & \small{ \space + KD + 8NN} & \small{54.7} & \small{+1.5} \\
 \small{7} & \small{ \space + KD + 8NN + reranked (=vanilla)} & \small{55.2} & \small{+0.5} \\
 \small{8} & \small{ \space + \nkd{8} + $\epsilon$} & \small{55.0} & \small{-0.2} \\
 \small{9} & \small{ \space + \nmm{8} + $\epsilon$} & \small{55.4} & \small{+0.4} \\
 \hline
\end{tabular}
\caption{\label{tab:ablation} Ablation study of {\minimax} on SST-5. $\Delta$ denotes the performance difference with respect to the previous row, but for the first row in each section, it indicates the difference with the accuracy of KD---i.e., row 2.}
\end{table}

\subsection{Ablation Study} \label{sec:ablation}

We analyze each component of our augmentation strategy to understand how they impact the overall effectiveness of {\minimax}. To this end, three components of our strategy are targeted for an ablation study. First, the effect of nearest neighbours is measured by replacing them with random examples from the sentence repository. Then, to determine whether reranking neighbours by teacher is helpful, we preserve the order of nearest neighbours returned by the SASE. Finally, we relax the maximum radial distance to include all nearest neighbours.

In Table~\ref{tab:ablation}, we report the results on SST-5. Surprisingly, random augmentation (row 3) scores only 0.3\% lower than {\knn} augmentation (row 6). Reranking nearest neighbours by the teacher further boosts the results by 0.5\% (row 7). The presence of maximum radial distance is not helpful for {\vanilla} as it leads to 0.4\% drop in the accuracy (row 8). Finally, our selection mechanism in {\minimax} (row 9) leads to a 0.2\% improvement compared to {\vanilla} (row 7).

\begin{table*}[t]
\centering
\small
 \begin{tabular}{l c} 
 \hline
 \rowcolor{green!8} \small{(i) \textbf{SST-5:} this is a stunning film, a one-of-a-kind tour de force.} & \small{{\bf very positive}} \\ 
 \rowcolor{green!8} \footnotesize{\space\space\space\space Here is masterful film-making in action. (5)} & \small{{\em very positive}} \\
 \rowcolor{green!8} \footnotesize{\space\space\space\space It's an expertly-crafted spectacle-event movie. (1)} & \small{{\em very positive}} \\
 \rowcolor{green!8} \footnotesize{\space\space\space\space This is a unique cinematographic experience. (6)} & \small{{\em very positive}} \\
 \hline
 \rowcolor{green!8} \small{(ii) \textbf{CR:} one also exhibited extremely slow speed when going to the menu.} & \small{{\bf negative}} \\
 \rowcolor{green!8} \footnotesize{\space\space\space\space No menu appears to make it very quick and easy to use. (15)} & \small{{\em negative}} \\
 \rowcolor{green!8} \footnotesize{\space\space\space\space Switching between options in the main menu is relatively slow. (13)} & \small{{\em negative}} \\
 \rowcolor{green!8} \footnotesize{\space\space\space\space the only niggle i have found is that the menus are a bit slow at times. (8)} & \small{{\em negative}} \\
 \hline
 \rowcolor{gray!20} \small{(iii) \textbf{SST-5:} final verdict: you've seen it all before.} & \small{{\bf very negative}} \\
 \rowcolor{gray!20} \footnotesize{\space\space\space\space Below is the final result. (15)} & \small{{\em neutral}} \\
 \rowcolor{gray!20} \footnotesize{\space\space\space\space The final verdict: Go ahead and buy (4)} & \small{{\em positive}} \\
 \rowcolor{gray!20} \footnotesize{\space\space\space\space Nut in the end, the final result always pays out. (7)} & \small{{\em positive}} \\
 \hline
 \rowcolor{red!10} \small{(iv) \textbf{TREC:} What causes the body to shiver in cold temperatures?} & \small{{\bf DESC}} \\
 \rowcolor{red!10} \footnotesize{\space\space\space\space}\scriptsize{How is it possible that a higher minimum wage could actually lead to more inequality within a country? (11)} & \small{{\em DESC}} \\
\rowcolor{red!10} \footnotesize{\space\space\space\space How did the minimum wage increase come about? (13)} & \small{{\em DESC}} \\
\rowcolor{red!10} \footnotesize{\space\space\space\space How is the new minimum wage hike impacting them? (7)} & \small{{\em DESC}} \\
 \hline
\end{tabular}
\caption{\label{tab:examples} Examples, derived from the augmented CR, SST-5, and TREC, after teacher reranking (the numbers in the bracket indicate the initial rank by SASE). For the nearest neighbours, the teacher's {\em predictions} are also provided, although soft labels will be used during training. Row (iii) shows an example of label mismatch and row (iv) highlights a mediocre paraphrase retrieval despite matching labels.}
\end{table*}

\subsection{Few-shot experiments}
\label{sec:fewshot}
Our data augmentation strategy can be applied to few-shot learning scenarios where a minuscule number of labelled data is available. Therefore, we simulate a few-shot learning setting as described in Section~\ref{ssec:setup}. In addition to {\vanilla} and no augmentation baselines, we compare our results with SentAugment \cite{du-etal-2021-self}, the state-of-the-art method in {\knn} data augmentation. In SentAugment, experiments are conducted on 5 randomly sampled small datasets and top 3 results of 10 different runs are averaged across sampled datasets, which means average over 15 runs in total. To be comparable to SentAugment, we average across all 10 runs for 2 sampled datasets, average over 20 runs in total, to report our results. In SentAugment, augmented few-shot datasets contain 1000 examples including the original data. For {\minimax}, we use 10-NNs in this experiment with a maximum radial distance.

Table~\ref{tab:fewshot} shows the few-shot learning results. The performance of our baselines follows a similar trend in the full-size data experiments. In particular, KD without augmentation slightly improves the test accuracy; {\Vanilla} brings almost 1.9\% improvement on average, and {\minimax} consistently surpasses {\vanilla} by 0.7\% on average. Compared to SentAugment, {\minimax} reaches a competitive performance. The key advantage of {\minimax} lies in sample efficiency. Specifically, {\minimax} falls short by only 0.3\% on SST-2, and IMP with using less than 40\% of the SentAugment augmented data on average. On CR, {\minimax} lags behind by 1.4\%, but again on roughly 40\% of the SentAugment data size. Moreover, SentAugment outperforms our approach by 0.3\% on TREC, while the size of augmentation is reduced by almost 20\%. Lastly, {\minimax} outperforms SentAugment in SST-5 by 2.4\% with almost same amount of data.

\subsection{Qualitative Analysis}
We study the quality of augmented examples retrieved from the sentence repository. Table~\ref{tab:examples} presents four examples from SST-5, CR, and TREC along with the corresponding top 3-NNs. The top two rows show clear-cut examples that the nearest neighbours are in fact paraphrased forms of original samples. Also, the teacher predicts the same label as the original examples for these augmented examples. We observe task-specific knowledge that the teacher has learned from original data helps to rank retrieved sentences---e.g., in the second row, reranking pushes the neighbours at ranks 15 and 13 to the top 3.

However, the augmented data is not always perfect. To identify the limitations of {\knn} augmentation, we manually inspect 20 samples, randomly drawn from SST-5 and TREC. We find that inaccurate paraphrase retrieval undercuts the quality of augmented examples shown in the bottom rows of Table~\ref{tab:examples}. A side effect of this weakness is the domain mismatch, denoting that augmentation can introduce out-of-domain data. For instance, the input data in TREC is expected to be in interrogative mood, but the retrieval may return declarative sentences. A potential solution to this problem could be utilizing a different repository, entirely comprised of questions in this case, similar to that of \citet{perez-etal-2020-unsupervised}. However, curating such repository for specialized domains can be challenging. Moreover, the improvements we observe in the experiments show that this issue is not prevalent in our selected tasks.

\section{Conclusion}
In this paper, we presented a sample efficient semi-supervised data augmentation technique, namely {\minimax}. The augmentation procedure is framed as finding nearest neighbours from a massive repository of unannotated sentences. The crucial aspect of {\knn} augmentation is interpretability as augmented examples are written in natural language. We adopt KD to learn from unlabelled data. The key ingredient of our approach is to find the most impactful examples that maximize the KL-divergence between the teacher and the student models. We show that {\minimax} can reduce the augmented data size by 50\% while improving upon vanilla augmentation.

\section*{Acknowledgments}
We thank MindSpore\footnote{\url{https://www.mindspore.cn}}---a new deep learning framework---for partially supporting this work.

\bibliographystyle{acl_natbib}
\bibliography{anthology,acl2021}

\begin{thebibliography}{53}
\expandafter\ifx\csname natexlab\endcsname\relax\def\natexlab#1{#1}\fi

\bibitem[{Buciluǎ et~al.(2006)Buciluǎ, Caruana, and
  Niculescu-Mizil}]{bucilua2006model}
Cristian Buciluǎ, Rich Caruana, and Alexandru Niculescu-Mizil. 2006.
\newblock \href {https://doi.org/10.1145/1150402.1150464} {Model compression}.
\newblock In \emph{Proceedings of the 12th ACM SIGKDD International Conference
  on Knowledge Discovery and Data Mining}, KDD '06, pages 535--541, New York,
  NY, USA. Association for Computing Machinery.

\bibitem[{Chen et~al.(2020{\natexlab{a}})Chen, Wu, and Yang}]{chen2020semi}
Jiaao Chen, Yuwei Wu, and Diyi Yang. 2020{\natexlab{a}}.
\newblock \href {http://arxiv.org/abs/2004.10972} {Semi-supervised models via
  data augmentation for classifying interactive affective responses}.
\newblock In \emph{AffCon@ AAAI}.

\bibitem[{Chen et~al.(2020{\natexlab{b}})Chen, Kornblith, Swersky, Norouzi, and
  Hinton}]{chen2020big}
Ting Chen, Simon Kornblith, Kevin Swersky, Mohammad Norouzi, and Geoffrey~E
  Hinton. 2020{\natexlab{b}}.
\newblock \href
  {https://proceedings.neurips.cc/paper/2020/file/fcbc95ccdd551da181207c0c1400c655-Paper.pdf}
  {Big self-supervised models are strong semi-supervised learners}.
\newblock In \emph{Advances in Neural Information Processing Systems},
  volume~33, pages 22243--22255. Curran Associates, Inc.

\bibitem[{Cheng et~al.(2020)Cheng, Jiang, Macherey, and
  Eisenstein}]{cheng-etal-2020-advaug}
Yong Cheng, Lu~Jiang, Wolfgang Macherey, and Jacob Eisenstein. 2020.
\newblock \href {https://doi.org/10.18653/v1/2020.acl-main.529} {{A}dv{A}ug:
  Robust adversarial augmentation for neural machine translation}.
\newblock In \emph{Proceedings of the 58th Annual Meeting of the Association
  for Computational Linguistics}, pages 5961--5970, Online. Association for
  Computational Linguistics.

\bibitem[{Clark et~al.(2019)Clark, Luong, Khandelwal, Manning, and
  Le}]{clark-etal-2019-bam}
Kevin Clark, Minh-Thang Luong, Urvashi Khandelwal, Christopher~D. Manning, and
  Quoc~V. Le. 2019.
\newblock \href {https://doi.org/10.18653/v1/P19-1595} {{BAM}! born-again
  multi-task networks for natural language understanding}.
\newblock In \emph{Proceedings of the 57th Annual Meeting of the Association
  for Computational Linguistics}, pages 5931--5937, Florence, Italy.
  Association for Computational Linguistics.

\bibitem[{Clark et~al.(2020)Clark, Luong, Le, and Manning}]{clark2020electra}
Kevin Clark, Minh-Thang Luong, Quoc~V. Le, and Christopher~D. Manning. 2020.
\newblock \href {https://openreview.net/forum?id=r1xMH1BtvB} {{ELECTRA}:
  Pre-training text encoders as discriminators rather than generators}.
\newblock In \emph{International Conference on Learning Representations}.

\bibitem[{Du et~al.(2021)Du, Grave, Gunel, Chaudhary, Celebi, Auli, Stoyanov,
  and Conneau}]{du-etal-2021-self}
Jingfei Du, Edouard Grave, Beliz Gunel, Vishrav Chaudhary, Onur Celebi, Michael
  Auli, Veselin Stoyanov, and Alexis Conneau. 2021.
\newblock \href {https://www.aclweb.org/anthology/2021.naacl-main.426}
  {Self-training improves pre-training for natural language understanding}.
\newblock In \emph{Proceedings of the 2021 Conference of the North American
  Chapter of the Association for Computational Linguistics: Human Language
  Technologies}, pages 5408--5418, Online. Association for Computational
  Linguistics.

\bibitem[{Fu et~al.(2020)Fu, Geng, Duan, Zhuang, Yuan, Trischler, Lin, Pal, and
  Dong}]{fu2020role}
Jie Fu, Xue Geng, Zhijian Duan, Bohan Zhuang, Xingdi Yuan, Adam Trischler, Jie
  Lin, Chris Pal, and Hao Dong. 2020.
\newblock \href {http://arxiv.org/abs/2004.08861} {Role-wise data augmentation
  for knowledge distillation}.
\newblock arXiv:2004.08861.

\bibitem[{Furlanello et~al.(2018)Furlanello, Lipton, Tschannen, Itti, and
  Anandkumar}]{furlanello2018born}
Tommaso Furlanello, Zachary Lipton, Michael Tschannen, Laurent Itti, and Anima
  Anandkumar. 2018.
\newblock \href {http://proceedings.mlr.press/v80/furlanello18a.html} {Born
  again neural networks}.
\newblock In \emph{Proceedings of the 35th International Conference on Machine
  Learning}, volume~80, pages 1607--1616. PMLR.

\bibitem[{Guu et~al.(2020)Guu, Lee, Tung, Pasupat, and Chang}]{guu2020realm}
Kelvin Guu, Kenton Lee, Zora Tung, Panupong Pasupat, and Ming-Wei Chang. 2020.
\newblock \href {http://arxiv.org/abs/2002.08909} {Retrieval-augmented language
  model pre-training}.
\newblock In \emph{Proceedings of the 37th International Conference on Machine
  Learning}, volume 119, pages 3929--3938.

\bibitem[{Hinton et~al.(2015)Hinton, Vinyals, and Dean}]{hinton2015distilling}
Geoffrey Hinton, Oriol Vinyals, and Jeff Dean. 2015.
\newblock \href {https://arxiv.org/abs/1503.02531} {Distilling the knowledge in
  a neural network}.
\newblock arXiv:1503.02531.

\bibitem[{Hu and Liu(2004)}]{hu2004mining}
Minqing Hu and Bing Liu. 2004.
\newblock \href {https://doi.org/10.1145/1014052.1014073} {Mining and
  summarizing customer reviews}.
\newblock In \emph{Proceedings of the Tenth ACM SIGKDD International Conference
  on Knowledge Discovery and Data Mining}, KDD '04, pages 168--177.

\bibitem[{Jiang et~al.(2020)Jiang, He, Chen, Liu, Gao, and
  Zhao}]{jiang-etal-2020-smart}
Haoming Jiang, Pengcheng He, Weizhu Chen, Xiaodong Liu, Jianfeng Gao, and Tuo
  Zhao. 2020.
\newblock \href {https://doi.org/10.18653/v1/2020.acl-main.197} {{SMART}:
  Robust and efficient fine-tuning for pre-trained natural language models
  through principled regularized optimization}.
\newblock In \emph{Proceedings of the 58th Annual Meeting of the Association
  for Computational Linguistics}, pages 2177--2190, Online. Association for
  Computational Linguistics.

\bibitem[{Jiao et~al.(2020)Jiao, Yin, Shang, Jiang, Chen, Li, Wang, and
  Liu}]{jiao-etal-2020-tinybert}
Xiaoqi Jiao, Yichun Yin, Lifeng Shang, Xin Jiang, Xiao Chen, Linlin Li, Fang
  Wang, and Qun Liu. 2020.
\newblock \href {https://doi.org/10.18653/v1/2020.findings-emnlp.372}
  {{T}iny{BERT}: Distilling {BERT} for natural language understanding}.
\newblock In \emph{Findings of the Association for Computational Linguistics:
  EMNLP 2020}, pages 4163--4174, Online. Association for Computational
  Linguistics.

\bibitem[{Jin et~al.(2020)Jin, Jin, Zhou, and Szolovits}]{jin2020bert}
Di~Jin, Zhijing Jin, Joey~Tianyi Zhou, and Peter Szolovits. 2020.
\newblock \href {https://doi.org/10.1609/aaai.v34i05.6311} {Is {BERT} really
  robust? a strong baseline for natural language attack on text classification
  and entailment}.
\newblock In \emph{Proceedings of the AAAI Conference on Artificial
  Intelligence}, volume~34, pages 8018--8025.

\bibitem[{Kassner and Sch{\"u}tze(2020)}]{kassner-schutze-2020-bert}
Nora Kassner and Hinrich Sch{\"u}tze. 2020.
\newblock \href {https://doi.org/10.18653/v1/2020.findings-emnlp.307}
  {{BERT}-k{NN}: Adding a k{NN} search component to pretrained language models
  for better {QA}}.
\newblock In \emph{Findings of the Association for Computational Linguistics:
  EMNLP 2020}, pages 3424--3430, Online. Association for Computational
  Linguistics.

\bibitem[{Kaushik et~al.(2020)Kaushik, Hovy, and Lipton}]{kaushik2019learning}
Divyansh Kaushik, Eduard Hovy, and Zachary~C Lipton. 2020.
\newblock \href {https://openreview.net/forum?id=Sklgs0NFvr} {Learning the
  difference that makes a difference with counterfactually-augmented data}.
\newblock In \emph{International Conference on Learning Representations}.

\bibitem[{Khandelwal et~al.(2021)Khandelwal, Fan, Jurafsky, Zettlemoyer, and
  Lewis}]{khandelwal2021nearest}
Urvashi Khandelwal, Angela Fan, Dan Jurafsky, Luke Zettlemoyer, and Mike Lewis.
  2021.
\newblock \href {https://openreview.net/forum?id=7wCBOfJ8hJM} {Nearest neighbor
  machine translation}.
\newblock In \emph{International Conference on Learning Representations}.

\bibitem[{Khandelwal et~al.(2020)Khandelwal, Levy, Jurafsky, Zettlemoyer, and
  Lewis}]{khandelwal2019generalization}
Urvashi Khandelwal, Omer Levy, Dan Jurafsky, Luke Zettlemoyer, and Mike Lewis.
  2020.
\newblock \href {https://openreview.net/forum?id=HklBjCEKvH} {Generalization
  through memorization: Nearest neighbor language models}.
\newblock In \emph{International Conference on Learning Representations}.

\bibitem[{Khashabi et~al.(2020)Khashabi, Khot, and
  Sabharwal}]{khashabi-etal-2020-bang}
Daniel Khashabi, Tushar Khot, and Ashish Sabharwal. 2020.
\newblock \href {https://doi.org/10.18653/v1/2020.emnlp-main.12} {More bang for
  your buck: Natural perturbation for robust question answering}.
\newblock In \emph{Proceedings of the 2020 Conference on Empirical Methods in
  Natural Language Processing (EMNLP)}, pages 163--170, Online. Association for
  Computational Linguistics.

\bibitem[{Kobayashi(2018)}]{kobayashi2018contextual}
Sosuke Kobayashi. 2018.
\newblock \href {https://doi.org/10.18653/v1/N18-2072} {Contextual
  augmentation: Data augmentation by words with paradigmatic relations}.
\newblock In \emph{Proceedings of the 2018 Conference of the North {A}merican
  Chapter of the Association for Computational Linguistics: Human Language
  Technologies, Volume 2 (Short Papers)}, pages 452--457.

\bibitem[{Lample and Conneau(2019)}]{lample2019cross}
Guillaume Lample and Alexis Conneau. 2019.
\newblock \href
  {https://proceedings.neurips.cc/paper/2019/file/c04c19c2c2474dbf5f7ac4372c5b9af1-Paper.pdf}
  {Cross-lingual language model pretraining}.
\newblock In \emph{Advances in Neural Information Processing Systems},
  volume~32, pages 7059--7069.

\bibitem[{Lewis et~al.(2020)Lewis, Perez, Piktus, Petroni, Karpukhin, Goyal,
  K\"{u}ttler, Lewis, Yih, Rockt\"{a}schel, Riedel, and
  Kiela}]{lewis2020retrieval}
Patrick Lewis, Ethan Perez, Aleksandra Piktus, Fabio Petroni, Vladimir
  Karpukhin, Naman Goyal, Heinrich K\"{u}ttler, Mike Lewis, Wen-tau Yih, Tim
  Rockt\"{a}schel, Sebastian Riedel, and Douwe Kiela. 2020.
\newblock \href
  {https://proceedings.neurips.cc/paper/2020/file/6b493230205f780e1bc26945df7481e5-Paper.pdf}
  {Retrieval-augmented generation for knowledge-intensive {NLP} tasks}.
\newblock In \emph{Advances in Neural Information Processing Systems},
  volume~33, pages 9459--9474. Curran Associates, Inc.

\bibitem[{Li and Roth(2002)}]{li2002learning}
Xin Li and Dan Roth. 2002.
\newblock \href {https://www.aclweb.org/anthology/C02-1150} {Learning question
  classifiers}.
\newblock In \emph{{COLING} 2002: The 19th International Conference on
  Computational Linguistics}.

\bibitem[{Liu et~al.(2019)Liu, Ott, Goyal, Du, Joshi, Chen, Levy, Lewis,
  Zettlemoyer, and Stoyanov}]{liu2019roberta}
Yinhan Liu, Myle Ott, Naman Goyal, Jingfei Du, Mandar Joshi, Danqi Chen, Omer
  Levy, Mike Lewis, Luke Zettlemoyer, and Veselin Stoyanov. 2019.
\newblock \href {http://arxiv.org/abs/1907.11692} {{RoBERTa}: A robustly
  optimized {BERT} pretraining approach}.
\newblock arXiv:1907.11692.

\bibitem[{Lopez-Paz et~al.(2015)Lopez-Paz, Bottou, Sch{\"o}lkopf, and
  Vapnik}]{lopez2015unifying}
David Lopez-Paz, L{\'e}on Bottou, Bernhard Sch{\"o}lkopf, and Vladimir Vapnik.
  2015.
\newblock \href {http://arxiv.org/abs/1511.03643} {Unifying distillation and
  privileged information}.
\newblock arXiv:1511.03643.

\bibitem[{Madry et~al.(2018)Madry, Makelov, Schmidt, Tsipras, and
  Vladu}]{madry2017towards}
Aleksander Madry, Aleksandar Makelov, Ludwig Schmidt, Dimitris Tsipras, and
  Adrian Vladu. 2018.
\newblock \href {https://openreview.net/forum?id=rJzIBfZAb} {Towards deep
  learning models resistant to adversarial attacks}.
\newblock In \emph{International Conference on Learning Representations}.

\bibitem[{Miyato et~al.(2017)Miyato, Dai, and
  Goodfellow}]{miyato2017adversarial}
Takeru Miyato, Andrew~M Dai, and Ian Goodfellow. 2017.
\newblock \href {http://arxiv.org/abs/1605.07725} {Adversarial training methods
  for semi-supervised text classification}.

\bibitem[{Mosbach et~al.(2021)Mosbach, Andriushchenko, and
  Klakow}]{mosbach2020stability}
Marius Mosbach, Maksym Andriushchenko, and Dietrich Klakow. 2021.
\newblock \href {https://openreview.net/forum?id=nzpLWnVAyah} {On the stability
  of fine-tuning {BERT}: Misconceptions, explanations, and strong baselines}.
\newblock In \emph{International Conference on Learning Representations}.

\bibitem[{Nayak et~al.(2019)Nayak, Mopuri, Shaj, Radhakrishnan, and
  Chakraborty}]{nayak2019zero}
Gaurav~Kumar Nayak, Konda~Reddy Mopuri, Vaisakh Shaj, Venkatesh~Babu
  Radhakrishnan, and Anirban Chakraborty. 2019.
\newblock \href {http://proceedings.mlr.press/v97/nayak19a.html} {Zero-shot
  knowledge distillation in deep networks}.
\newblock In \emph{Proceedings of the 36th International Conference on Machine
  Learning}, volume~97, pages 4743--4751. PMLR.

\bibitem[{Ng et~al.(2020)Ng, Cho, and Ghassemi}]{ng-etal-2020-ssmba}
Nathan Ng, Kyunghyun Cho, and Marzyeh Ghassemi. 2020.
\newblock \href {https://doi.org/10.18653/v1/2020.emnlp-main.97} {{SSMBA}:
  Self-supervised manifold based data augmentation for improving out-of-domain
  robustness}.
\newblock In \emph{Proceedings of the 2020 Conference on Empirical Methods in
  Natural Language Processing (EMNLP)}, pages 1268--1283, Online. Association
  for Computational Linguistics.

\bibitem[{Passban et~al.(2020)Passban, Wu, Rezagholizadeh, and
  Liu}]{passban2020alp}
Peyman Passban, Yimeng Wu, Mehdi Rezagholizadeh, and Qun Liu. 2020.
\newblock \href {http://arxiv.org/abs/2012.14022} {{ALP-KD}: Attention-based
  layer projection for knowledge distillation}.
\newblock arXiv:2012.14022.

\bibitem[{Perez et~al.(2020)Perez, Lewis, Yih, Cho, and
  Kiela}]{perez-etal-2020-unsupervised}
Ethan Perez, Patrick Lewis, Wen-tau Yih, Kyunghyun Cho, and Douwe Kiela. 2020.
\newblock \href {https://doi.org/10.18653/v1/2020.emnlp-main.713} {Unsupervised
  question decomposition for question answering}.
\newblock In \emph{Proceedings of the 2020 Conference on Empirical Methods in
  Natural Language Processing (EMNLP)}, pages 8864--8880, Online. Association
  for Computational Linguistics.

\bibitem[{Qu et~al.(2021)Qu, Shen, Shen, Sajeev, Han, and Chen}]{qu2021coda}
Yanru Qu, Dinghan Shen, Yelong Shen, Sandra Sajeev, Jiawei Han, and Weizhu
  Chen. 2021.
\newblock \href {https://openreview.net/forum?id=Ozk9MrX1hvA} {{CoDA}:
  Contrast-enhanced and diversity-promoting data augmentation for natural
  language understanding}.
\newblock In \emph{International Conference on Learning Representations}.

\bibitem[{Rashid et~al.(2020)Rashid, Lioutas, Ghaddar, and
  Rezagholizadeh}]{rashid2020towards}
Ahmad Rashid, Vasileios Lioutas, Abbas Ghaddar, and Mehdi Rezagholizadeh. 2020.
\newblock \href {http://arxiv.org/abs/2012.15495} {Towards zero-shot knowledge
  distillation for natural language processing}.
\newblock arXiv:2012.15495.

\bibitem[{Rashid et~al.(2021)Rashid, Lioutas, and
  Rezagholizadeh}]{rashid2021mate}
Ahmad Rashid, Vasileios Lioutas, and Mehdi Rezagholizadeh. 2021.
\newblock \href {http://arxiv.org/abs/2105.05912} {{MATE-KD}: {Masked
  Adversarial TExt}, a companion to knowledge distillation}.
\newblock arXiv:2105.05912.

\bibitem[{Sanh et~al.(2019)Sanh, Debut, Chaumond, and
  Wolf}]{sanh2019distilbert}
Victor Sanh, Lysandre Debut, Julien Chaumond, and Thomas Wolf. 2019.
\newblock \href {http://arxiv.org/abs/1910.01108} {{DistilBERT}, a distilled
  version of {BERT}: smaller, faster, cheaper and lighter}.
\newblock arXiv:1910.01108.

\bibitem[{Sennrich et~al.(2016)Sennrich, Haddow, and
  Birch}]{sennrich2016improving}
Rico Sennrich, Barry Haddow, and Alexandra Birch. 2016.
\newblock \href {https://www.aclweb.org/anthology/P16-1009} {Improving neural
  machine translation models with monolingual data}.
\newblock In \emph{Proceedings of the 54th Annual Meeting of the Association
  for Computational Linguistics (Volume 1: Long Papers)}, pages 86--96, Berlin,
  Germany. Association for Computational Linguistics.

\bibitem[{Socher et~al.(2013)Socher, Perelygin, Wu, Chuang, Manning, Ng, and
  Potts}]{socher2013recursive}
Richard Socher, Alex Perelygin, Jean Wu, Jason Chuang, Christopher~D Manning,
  Andrew~Y Ng, and Christopher Potts. 2013.
\newblock \href {https://www.aclweb.org/anthology/D13-1170} {Recursive deep
  models for semantic compositionality over a sentiment treebank}.
\newblock In \emph{Proceedings of the 2013 Conference on Empirical Methods in
  Natural Language Processing}, pages 1631--1642, Seattle, Washington, USA.
  Association for Computational Linguistics.

\bibitem[{Tan et~al.(2019)Tan, Ren, He, Qin, Zhao, and
  Liu}]{tan2019multilingual}
Xu~Tan, Yi~Ren, Di~He, Tao Qin, Zhou Zhao, and Tie-Yan Liu. 2019.
\newblock \href {https://openreview.net/forum?id=S1gUsoR9YX} {Multilingual
  neural machine translation with knowledge distillation}.
\newblock In \emph{International Conference on Learning Representations}.

\bibitem[{Turc et~al.(2019)Turc, Chang, Lee, and Toutanova}]{turc2019well}
Iulia Turc, Ming-Wei Chang, Kenton Lee, and Kristina Toutanova. 2019.
\newblock \href {http://arxiv.org/abs/1908.08962} {Well-read students learn
  better: On the importance of pre-training compact models}.
\newblock arXiv:1908.08962.

\bibitem[{Volpi et~al.(2018)Volpi, Namkoong, Sener, Duchi, Murino, and
  Savarese}]{volpi2018generalizing}
Riccardo Volpi, Hongseok Namkoong, Ozan Sener, John~C Duchi, Vittorio Murino,
  and Silvio Savarese. 2018.
\newblock \href
  {https://proceedings.neurips.cc/paper/2018/file/1d94108e907bb8311d8802b48fd54b4a-Paper.pdf}
  {Generalizing to unseen domains via adversarial data augmentation}.
\newblock In \emph{Advances in Neural Information Processing Systems},
  volume~31. Curran Associates, Inc.

\bibitem[{Wei and Zou(2019)}]{wei-zou-2019-eda}
Jason Wei and Kai Zou. 2019.
\newblock \href {https://doi.org/10.18653/v1/D19-1670} {{EDA}: Easy data
  augmentation techniques for boosting performance on text classification
  tasks}.
\newblock In \emph{Proceedings of the 2019 Conference on Empirical Methods in
  Natural Language Processing and the 9th International Joint Conference on
  Natural Language Processing (EMNLP-IJCNLP)}, pages 6382--6388, Hong Kong,
  China. Association for Computational Linguistics.

\bibitem[{Wolf et~al.(2020)Wolf, Debut, Sanh, Chaumond, Delangue, Moi, Cistac,
  Rault, Louf, Funtowicz, Davison, Shleifer, von Platen, Ma, Jernite, Plu, Xu,
  Le~Scao, Gugger, Drame, Lhoest, and Rush}]{wolf-etal-2020-transformers}
Thomas Wolf, Lysandre Debut, Victor Sanh, Julien Chaumond, Clement Delangue,
  Anthony Moi, Pierric Cistac, Tim Rault, Remi Louf, Morgan Funtowicz, Joe
  Davison, Sam Shleifer, Patrick von Platen, Clara Ma, Yacine Jernite, Julien
  Plu, Canwen Xu, Teven Le~Scao, Sylvain Gugger, Mariama Drame, Quentin Lhoest,
  and Alexander Rush. 2020.
\newblock \href {https://doi.org/10.18653/v1/2020.emnlp-demos.6} {Transformers:
  State-of-the-art natural language processing}.
\newblock In \emph{Proceedings of the 2020 Conference on Empirical Methods in
  Natural Language Processing: System Demonstrations}, pages 38--45, Online.
  Association for Computational Linguistics.

\bibitem[{Wu et~al.(2020)Wu, Passban, Rezagholizadeh, and
  Liu}]{wu-etal-2020-skip}
Yimeng Wu, Peyman Passban, Mehdi Rezagholizadeh, and Qun Liu. 2020.
\newblock \href {https://doi.org/10.18653/v1/2020.emnlp-main.74} {Why skip if
  you can combine: A simple knowledge distillation technique for intermediate
  layers}.
\newblock In \emph{Proceedings of the 2020 Conference on Empirical Methods in
  Natural Language Processing (EMNLP)}, pages 1016--1021, Online. Association
  for Computational Linguistics.

\bibitem[{Xie et~al.(2020{\natexlab{a}})Xie, Dai, Hovy, Luong, and
  Le}]{xie2020unsupervised}
Qizhe Xie, Zihang Dai, Eduard Hovy, Thang Luong, and Quoc Le.
  2020{\natexlab{a}}.
\newblock \href
  {https://proceedings.neurips.cc/paper/2020/file/44feb0096faa8326192570788b38c1d1-Paper.pdf}
  {Unsupervised data augmentation for consistency training}.
\newblock In \emph{Advances in Neural Information Processing Systems},
  volume~33, pages 6256--6268. Curran Associates, Inc.

\bibitem[{Xie et~al.(2020{\natexlab{b}})Xie, Luong, Hovy, and Le}]{xie2020self}
Qizhe Xie, Minh-Thang Luong, Eduard Hovy, and Quoc~V Le. 2020{\natexlab{b}}.
\newblock \href {https://arxiv.org/abs/1911.04252} {Self-training with noisy
  student improves {ImageNet} classification}.
\newblock In \emph{Proceedings of the IEEE/CVF Conference on Computer Vision
  and Pattern Recognition (CVPR)}, pages 10687--10698.

\bibitem[{Xie et~al.(2017)Xie, Wang, Li, L{\'e}vy, Nie, Jurafsky, and
  Ng}]{xie2017data}
Ziang Xie, Sida~I Wang, Jiwei Li, Daniel L{\'e}vy, Aiming Nie, Dan Jurafsky,
  and Andrew~Y Ng. 2017.
\newblock \href {https://openreview.net/forum?id=H1VyHY9gg} {Data noising as
  smoothing in neural network language models}.
\newblock In \emph{International Conference on Learning Representations}.

\bibitem[{Yi et~al.(2021)Yi, Hou, Shang, Jiang, Liu, and
  Ma}]{yi2021reweighting}
Mingyang Yi, Lu~Hou, Lifeng Shang, Xin Jiang, Qun Liu, and Zhi-Ming Ma. 2021.
\newblock \href {https://openreview.net/forum?id=9G5MIc-goqB} {Reweighting
  augmented samples by minimizing the maximal expected loss}.
\newblock In \emph{International Conference on Learning Representations}.

\bibitem[{Yu et~al.(2018)Yu, Dohan, Luong, Zhao, Chen, Norouzi, and
  Le}]{yu2018qanet}
Adams~Wei Yu, David Dohan, Minh-Thang Luong, Rui Zhao, Kai Chen, Mohammad
  Norouzi, and Quoc~V Le. 2018.
\newblock \href {https://openreview.net/forum?id=B14TlG-RW} {{QANet}: Combining
  local convolution with global self-attention for reading comprehension}.
\newblock In \emph{International Conference on Learning Representations}.

\bibitem[{Zhang et~al.(2021)Zhang, Wu, Katiyar, Weinberger, and
  Artzi}]{zhang2021revisiting}
Tianyi Zhang, Felix Wu, Arzoo Katiyar, Kilian~Q Weinberger, and Yoav Artzi.
  2021.
\newblock \href {https://openreview.net/forum?id=cO1IH43yUF} {Revisiting
  few-sample {BERT} fine-tuning}.
\newblock In \emph{International Conference on Learning Representations}.

\bibitem[{Zhang et~al.(2015)Zhang, Zhao, and LeCun}]{zhang2015character}
Xiang Zhang, Junbo Zhao, and Yann LeCun. 2015.
\newblock \href
  {https://proceedings.neurips.cc/paper/2015/file/250cf8b51c773f3f8dc8b4be867a9a02-Paper.pdf}
  {Character-level convolutional networks for text classification}.
\newblock In \emph{Advances in Neural Information Processing Systems},
  volume~28, pages 649--657. Curran Associates, Inc.

\bibitem[{Zhu et~al.(2020)Zhu, Cheng, Gan, Sun, Goldstein, and
  Liu}]{zhu2019freelb}
Chen Zhu, Yu~Cheng, Zhe Gan, Siqi Sun, Tom Goldstein, and Jingjing Liu. 2020.
\newblock \href {https://openreview.net/forum?id=BygzbyHFvB} {{FreeLB}:
  Enhanced adversarial training for natural language understanding}.
\newblock In \emph{International Conference on Learning Representations}.

\end{thebibliography}

\end{document}